\documentclass[letterpaper]{article} 
\usepackage{aaai24}  
\usepackage{times}  
\usepackage{helvet}  
\usepackage{courier}  
\usepackage[hyphens]{url}  
\usepackage{graphicx} 
\usepackage{natbib}  
\usepackage{caption} 
\usepackage{algorithm}
\usepackage{algorithmic}

\usepackage{amsthm}
\usepackage{amssymb}
\usepackage{amsmath}
\usepackage{graphicx}
\usepackage{booktabs}

\usepackage{newfloat}
\usepackage{listings}
\DeclareCaptionStyle{ruled}{labelfont=normalfont,labelsep=colon,strut=off} 
\floatstyle{ruled}
\newfloat{listing}{tb}{lst}{}
\floatname{listing}{Listing}
\usepackage{bibentry}
\usepackage[switch]{lineno}

\usepackage{color}

\title{DiffAIL: Diffusion Adversarial Imitation Learning}
\author{
    Bingzheng Wang\textsuperscript{\rm 1},
    Guoqiang Wu\textsuperscript{\rm 1}\textsuperscript{*},
    Teng Pang\textsuperscript{\rm 1},
    Yan Zhang\textsuperscript{\rm 1},
    Yilong Yin\textsuperscript{\rm 1}\textsuperscript{*}
}
\affiliations{
    \textsuperscript{\rm 1} Shandong University\\

    binzhwang@gmail.com, 
    guoqiangwu@sdu.edu.cn , \\
    silencept7@gmail.com, 
    yannzhang9@gmail.com,
    ylyin@sdu.edu.cn
}

\usepackage{bibentry}

\begin{document}

\maketitle

\begin{abstract}
Imitation learning aims to solve the problem of defining reward functions in real-world decision-making tasks. The current popular approach is the Adversarial Imitation Learning (AIL) framework, which matches expert state-action occupancy measures to obtain a surrogate reward for forward reinforcement learning. However, the traditional discriminator is a simple binary classifier and doesn't learn an accurate distribution, which may result in failing to identify expert-level state-action pairs induced by the policy interacting with the environment. To address this issue, we propose a method named diffusion adversarial imitation learning (DiffAIL), which introduces the diffusion model into the AIL framework. Specifically, DiffAIL models the state-action pairs as unconditional diffusion models and uses diffusion loss as part of the discriminator's learning objective, which enables the discriminator to capture better expert demonstrations and improve generalization. Experimentally, the results show that our method achieves state-of-the-art performance and significantly surpasses expert demonstration on two benchmark tasks, including the standard state-action setting and state-only settings. 
Our code can be available at the link https://github.com/ML-Group-SDU/DiffAIL.
\end{abstract}

\section{Introduction}

Deep Reinforcement Learning has achieved significant success in many decision-making tasks, including AlphaGo~\cite{silver2016mastering}, Atari games~\cite{mnih2013playing}, MuJoCo environment tasks~\cite{todorov2012mujoco}, and robot control tasks~\cite{mnih2015human}, where these tasks are typically defined with clear reward functions to guide the agent for decision-making. However, in real-world scenarios, the reward function is regularly challenging to obtain or define, such as in the case of autonomous driving, where it is complicated to delineate what behavior is beneficial for the agent. The emergence of \emph{Imitation Learning} provides a practical solution to the problem of inaccessible rewards. In imitation learning, the agent does not rely on actual rewards but instead utilizes expert demonstration data to learn a similar expert policy.

A relatively old method in imitation learning is clone learning~\cite{pomerleau1991efficient}, which uses supervised learning to learn from expert data. Although such methods are straightforward to implement, they are prone to serious extrapolation errors when visiting out-of-distribution data while practically interacting with environments. To alleviate the aforementioned errors, Dagger~\cite{ross2011reduction} proposes access to expert policies, where the agent continuously interacts with the environment online and asks for expert-level actions from the expert policy to expand the dataset. AdapMen~\cite{liu2023guide} proposes an active imitation learning framework based on teacher-student interaction, and theoretical analysis shows that it can avoid compounding errors under mild conditions.

To effectively tackle the aforementioned extrapolation error, Adversarial imitation learning (AIL)~\cite{ho2016generative,fu2017learning,kostrikov2018discriminator,ghasemipour2020divergence,zhang2020f,garg2021iq,zhang2022auto} has become the most popular approach in imitation learning. Rather than minimizing the divergence between the expert policy and the agent policy, AIL employs online interactive learning to focus on minimizing the divergence between the joint state-action distribution induced by the expert policy and the learned policy. However, typical AIL methods often use a simplistic discriminator that does not learn a distribution, which may not accurately classify specific expert-level actions generated by the policy during interactions with the environment. This can result in the naive discriminator failing to distinguish expert-level behaviors generated by the policy, potentially hindering agent learning. Therefore, a powerful discriminator that can accurately capture the distribution of expert data is crucial for AIL.

Currently, diffusion model~\cite{ho2020denoising} processes a powerful distribution matching capability and has achieved remarkable success in image generation, surpassing other popular generative models such as GAN~\cite{goodfellow2014generative} and VAE~\cite{kingma2013auto} in producing higher-quality and more diverse samples.

Based on diffusion model's powerful ability to capture data distribution, we propose a new method called Diffusion Adversarial Imitation Learning (DiffAIL) in this work. DiffAIL employs the same framework as traditional AIL but incorporates the diffusion model to model joint state-action distribution. Therefore, DiffAIL enables the discriminator to accurately capture expert demonstration and improve generalization, facilitating the successful identification of expert-level behaviors generated by the policy during interactions with the environment. Furthermore, we conducted experiments on representatively continuous control tasks in Mujoco, including the standard state-action setting and state-only setting. Surprisingly in both settings, we share the hyperparameters instead of readjusting them. The experiment results show that our method can achieve state-of-the-art (SOTA) performance and significantly surpass expert demonstration on these two benchmark tasks. 

Overall, our contributions can be summarized as follows:
\begin{itemize}
    \item We propose the DiffAIL method, which combines the diffusion model into AIL to improve the discriminator ability of distribution capturing.
    \item The experimental results show that DiffAIL can achieve SOTA performance, including standard state-action and state-only settings.
\end{itemize}

\section{Preliminaries}

\subsection{Problem Setting}
A sequential decision-making problem in reinforcement learning can be modeled as a Markov Decision Process (MDP) defined as a tuple $M = (S, A, \rho_0, P, r, \gamma)$, where $S$ denotes the state space, $A$ denotes the action space, $\rho_0$ denotes the initial state distribution, $P: S \times A \times S \rightarrow [0,1]$ describes the environment's dynamic model by specifying the state transition function, $r: S \times A \rightarrow \mathbb{R}$ denotes the reward function, and ${\gamma \in [0,1]}$ denotes the discount factor used to weigh the importance of future rewards. We represent the agent policy as $\pi: S \rightarrow A$. 
Policy $\pi(a|s)$ interacts with the environment to generate transitions $(s_t,a_t,r_t,s_{t+1})$,where $t$ denotes the timestep, $s_0\sim\rho_0, a_t\sim\pi(\cdot|s_t), r_t\sim r(s_t,a_t), s_{t+1}\sim p(\cdot|s_t,a_t)$. The goal of reinforcement learning is to learn a policy $\pi(a|s)$ maximizing cumulative discount rewards.

In imitation learning, the environment reward signal is not accessible. Rather, we can access expert demonstrations $D=\{(s_t, a_t)\}_{t=1}^{k}\sim\pi_e$ given by the unknown expert policy $\pi_e$. Imitation learning aims to learn a policy $\pi$  that can recover $\pi_e$  based on expert demonstrations without relying on reward signals. This setting can be purely offline to learn with only expert demonstrations or additionally online to interact with the environment (reward unavailable) by behavior policy

\subsection{Adversarial Imitation Learning}
For AIL, it aims to find an optimal policy that minimizes the state-action distribution divergence induced by the agent policy and the expert policy, as measured by the occupancy measure:
\begin{equation}
    \pi^*=\arg\min_\pi{D_f(d_\pi(s,a)||d_{\pi_e}(s,a))} ,
\end{equation}
where $d_\pi (s,a) = (1-\gamma)\sum\limits_{t=0}^{\infty} \gamma^t p(s_t=s,a_t=a)$ denotes state-action distribution of $\pi_\theta$, $(1-\gamma)$ term denotes normalization factor. Similarly, $d_{\pi_e}(s, a)$ denotes expert demonstration distribution with a similar form. $D_f$ can be an arbitrary distance formulation with the corresponding min-max target through the corresponding dual representation, which is also applied in f-gan~\cite{nowozin2016f}. Gail~\cite{ho2016generative} adopts the Jensen-Shannon (JS) divergence as the chosen distance metric $D_f$. By leveraging the dual representation of JS divergence, we can obtain the minimax optimization objective of Gail:
\begin{equation}
\begin{aligned}
    \min_{\pi_{\theta}} \max_{D_{\phi}} \ \mathbb{E}_{(s,a) \sim \pi_e} &[\log(D_{\phi}(s, a))]  \\
    & +\mathbb{E}_{(s,a) \sim \pi_{\theta}} [ \log(1 - D_{\phi}(s,a)) ] .
\end{aligned}
\end{equation}
The discriminator $D_\phi$ is used to identify between samples produced by the learning policy and expert demonstrations. Typically, the discriminator is a binary classifier whose output tends to be $1$ for expert data and $0$ for data generated by the policy.  The optimal discriminator satisfies the following:
\begin{equation}\label{eq3}
    \log{D^*(s,a)}-\log{(1-D^*(s,a))}=\log{\frac{d_{\pi_e}(s,a)}{d_{\pi_\theta}(s,a)}} ,
\end{equation}
where $D^*=\frac{d_{\pi_e} (s, a)}{d_{\pi_e} (s, a)+d_{\pi_\theta} (s, a)}$. The log-density ratio can be directly used as a surrogate reward function for forward reinforcement learning in CFIL~\cite{freund2023coupled}. Other popularly used surrogate reward function settings include $R(s,a) = \log{D_\phi(s,a)}$ or $R(s,a) = -\log{(1 - D_\phi(s,a))}$ used by Gail~\cite{ho2016generative} to minimize the Jensen-Shannon (JS) divergence and $R(s,a) = \log{D_\phi(s,a)} - \log{(1 - D_\phi(s,a))}$ used by AIRL~\cite{fu2017learning} to minimize the reverse Kullback-Leibler (KL) divergence. Although these surrogate reward functions have some prior bias in the absorbed state~\cite{kostrikov2018discriminator}, they still achieve good results in practical use.

All of these surrogate reward functions contain the discriminator term, which means that an advanced discriminator can provide better guidance for policy learning.

With these surrogate reward functions, Gail and other imitation methods can be combined with any forward reinforcement learning algorithm for policy optimization, such as PPO~\cite{schulman2017proximal} or TRPO~\cite{schulman2015trust} for on-policy algorithms and TD3~\cite{fujimoto2018addressing} or SAC~\cite{haarnoja2018soft} for off-policy algorithms.

\subsection{Diffusion Model}
Diffusion model~\cite{ho2020denoising} is a latent variable model mapped to potential space using a Markov chain. In the forward diffusion process, noise is gradually added to the data $x_t$ at each time step $t$ using a pre-defined variance schedule $\beta_t$. As $t$ increases, $x_t$ gets closer to pure noise, and when $T \rightarrow \infty$, $x_T$ becomes the standard Gaussian noise.
\begin{equation}
    q(x_{1:T}|x_0):=\prod_{t=1}^T{q(x_t|x_{t-1})} ,
\end{equation}
\begin{equation}\label{eq5}
    q(x_t|x_{t-1}):=\mathcal{N}(x_t;\sqrt{1-\beta_t}~x_{t-1},\beta_tI).
\end{equation}
By defining $\alpha_t=1-\beta_t$ and $\overline{\alpha}_t=\prod_{s=1}^{t}(1-\beta_s)$, we can use the reparameterization trick to directly sample $x_t$ at any time step $t$ from $x_0$:
\begin{equation}
    x_t=\sqrt{\overline{\alpha}_t} \ x_0 + \sqrt{1- \overline{\alpha}_t} \ \epsilon ,
\end{equation}
where $\epsilon \sim \mathcal{N}(0,1)$. The reverse diffusion process involves gradual sampling from the pure Gaussian noise to recover $x_0$. This process is modeled by latent variable models of the form $p_\phi(x_0):=\int p_\phi(x_{0:T})dx_{1:T}$.
\begin{equation}
    p_\phi(x_{0:T}):=p(x_T)\prod_{t=1}^{T} p_\phi (x_{t-1}|x_t) ,
\end{equation}
\begin{equation}
    p_\phi(x_{t-1}|x_t):=\mathcal{N}(x_{t-1}; ~ \mu_\phi(x_t, ~ t),{\Sigma}_\phi (x_t, ~ t)) ,
\end{equation}
where $\mu_\phi(x_t,t)=\frac{1}{\sqrt{1-\beta_t}}(x_t-\frac{\beta_t}{\sqrt{1-\overline{\alpha}_t}}\epsilon_\phi(x_t, ~ t))$ and $\epsilon_\phi$ is used to predict $\epsilon$ from $x_t$ in the reverse process by parameter $\phi$.
The covariance matrix ${\Sigma}_{\phi}(x_t,t)$ can be fixed for non-trainable or parameterized for learning.

The diffusion model maximizes the log-likelihood of the predicted distribution of the model, which is given by $\mathbb{L}=\mathbb{E}_{q(x_0)}[\log{p_\phi (x_0)}]$. Subsequently, similar to VAE~\cite{kingma2013auto}, the diffusion model maximizes the variational lower bound (VLB) defined as $\mathbb{E}_{q(x)}{[\ln{\frac{p_\phi (x_{0:T})}{q(x_{1:T}|x_0)}}]}$. The continuous derivation of VLB results in a simple loss function for the diffusion model, which can be expressed as follows:
\begin{align}
    L_{simple}(\phi)=\mathbb{E}_{ \ t \sim \mu(1,T), \ \epsilon\sim \mathcal{N}(0,I), \ x_0\sim{q(x_0)}}[||&\epsilon -  \notag \\
     \epsilon_\phi(\sqrt{ \overline\alpha_t} \ {x}_0 + \sqrt{1-\overline{\alpha}_t} \ \epsilon &, ~ t)||^{2}] ,
\end{align}
where $\overline{\alpha}_t=\prod_{s=1}^t{(1-\beta_s)}$, $\mu$ is a uniform distribution from between 1 and $T$ and $x_0$ sampled from real data.

\section{Method}
In this section, we present our proposed method - DiffAIL. First, we describe how we model the unconditional diffusion process on the joint distribution of state-action pairs. Second, we incorporate the diffusion model into the discriminator of an AIL framework to improve the ability that the discriminator to capture data distribution.

\subsection{Diffusion over state-action pairs}
To simplify the notation, we abuse notation and define $x_t = (s_i, a_i)_t$, where $t$ denotes the time step in the diffusion process and $i$ denotes the time at which a particular state is visited or an action in the trajectory. $x_t$ represents the state-action pairs available in the data. Since the forward diffusion process is parameter-free, we denote our forward diffusion process as follows:
\begin{equation}
    q(x_t|x_{t-1})=\mathcal{N}(x_t; ~ \sqrt{1-\beta_t} \ x_{t-1}, \beta_t I) ,
\end{equation}
where $\beta_t$ denotes the variance schedule at different time steps $t$. We then define the reverse process of the parameterized diffusion model as follows:
\begin{align}
    p_\phi(x_{0:T}|x_t)=\mathcal{N}(x_{t-1}; ~ \mu_\phi(x_t, ~ t), {\Sigma}_{\phi}(x_t, ~ t)) .
\end{align}
In our work, the covariance matrix is fixed as ${\Sigma}_{\phi} (x_t, ~ t) = \sigma_t^2 I = \beta_t I$ to be non-trainable and represented using a pre-defined time schedule $\beta_t$, following the same as in DDPM~\cite{ho2020denoising}. According to the forward process, $x_t$  can be derived from  $x_0$ at any time step $t$; the mean can be expressed as a noisy contained function:
\begin{equation}
    \begin{aligned}
        \mu_\phi(x_t, ~ t)=\frac{1}{\sqrt{1-\beta_t}}(x_t-\frac{\beta_t}{\sqrt{1-\overline{\alpha}_t}} \ \epsilon_\phi (x_t, ~ t)) .
    \end{aligned}
\end{equation}
Diffuser over state-action pairs is ultimately modelled as predicting the noise at each time step in the reverse diffusion process:
\begin{align}\label{eq13}
    L(\phi)=\mathbb{E}_{x_0 \sim D, \epsilon \sim \mathcal{N}(0,I), t \sim \mu(1,T)}  [Diff_\phi(x_0,\epsilon,t)], \\
    Diff_\phi(x_0,\epsilon,t)=||\epsilon - \epsilon_\phi(\sqrt{\overline{\alpha}_t}x_0+\sqrt{1-\overline{\alpha}_t} \ \epsilon, ~ t||^2 ,
\end{align}
where $\mu$ is the uniform distribution, $\epsilon$ follows the standard Gaussian noise distribution $\mathcal{N}(0, I)$ and $ \alpha_t= \prod_{s=1}^{T}(1-\beta_s) $.
In this work, we do not employ the diffusion model to infer for generating samples but use its loss as a distribution matching technique to improve the discriminator's ability to identify the expert demonstrations and policy data.
Therefore, practically,  $Diff_\phi$ can refer to either expert demonstrations $(s_i,a_i) \sim d_{\pi_e}(s, a)$  or data generated from interactions with the environment by policy $(s_i,a_i) \sim d_\pi(s, a)$.

Although many studies~\cite{song2020denoising,lu2022dpm,lu2022dpm++,zheng2023dpm} have focused on designing accelerated sampling algorithms for diffusion models to reduce the time cost of generating samples while maintaining high sample quality, we found that only a few tens of diffusion steps were sufficient to achieve excellent performance when modeling various environment tasks in MuJoCo, unlike image generation tasks that require thousands of sampling steps. Therefore, in our work, the primitive form of the diffusion model~\cite{ho2020denoising} is adequate and efficient.

\subsection{Diffusion Adversarial Imitation Learning}

\begin{table*}[htp]
\centering
\begin{tabular}{c|c|c|c|c}
\toprule
Task & Hopper & HalfCheetah & Ant & Walker2d \\
\midrule
Expert &3402 & 4463 & 4228 & 6717 \\
BC & 2376.22 ± 754.80 & 2781.85 ± 1143.11 &  2942.24 ± 344.62 & 1091.46 ± 187.60 \\ 
Gail &  2915.67 ± 328.12 & 4383.44 ± 62.08 & 4530.43 ± 133.30 & 1942.81 ± 957.68 \\ 
Valuedice &  2455.25 ± 658.43 & 4734.59 ± 229.67 & 3781.56 ± 571.15 & 4606.50 ± 927.80 \\ 
CFIL &  3234.61 ± 266.62 & 4976.73 ± 60.41 &  4222.07 ± 201.32 & 5039.77 ± 307.17 \\ 
\midrule
\textbf{DiffAIL}& \textbf{3382.03 ± 142.86} & \textbf{5362.25 ± 96.92} & \textbf{5142.60 ± 90.05} & \textbf{6292.28 ± 97.65} \\ 
\bottomrule
\end{tabular}
\caption{Learned policy best performance during training for different sota imitation learning algorithms with 1 trajectory over 5 seeds in the standard state-action setting.}
\label{table_1_traj}
\end{table*}

Next, we describe how diffusion loss is integrated into the AIL framework to improve discriminator generalization.

We introduce the diffusion model into adversarial imitation learning, which combines the simple loss function of the diffusion model with the traditional discriminator structure of AIL.
For convenience, we abbreviate $x_0$ as $x$ in the following text. The objective function used in our method is the same as Gail~\cite{ho2016generative}:
\begin{align}\label{eq15}
    \min_{{\pi_\theta}}\max_{D_{\phi}} \mathbb{E}_{ \epsilon \sim \mathcal{N}(0,I), t \sim \mu(1,T)} [&\mathbb{E}_{x \sim {\pi_e} }[\log(D_{\phi}(x,\epsilon,t))] \notag  \\ 
      + ~ \mathbb{E}_{x \sim \pi_\theta}[&\log(1-D_{\phi}(x,\epsilon,t))]],
\end{align}
where the $D_\phi(x,\epsilon,t)$ is a discrimination composed of diffusion noise loss by parameter $\phi$, and the policy network’s parameters are represented as $\theta$. At the inner level maximization discriminator provides higher values for expert data demonstrations and penalizes all other data areas. The specific formulation of $D_\phi(x,\epsilon,t)$ is defined as follows: 
\begin{align}\label{eq16}
    D_\phi(x,\epsilon,t)&=\exp{(-Diff_\phi(x,\epsilon,t))} \\ \notag
    &=\exp{(-||\epsilon - \epsilon_\phi(\sqrt{\overline\alpha_i} \ x +\sqrt{1-   \overline{\alpha}_t} \ \epsilon, ~ t)||^{2})} .
\end{align}
We apply the exponential operation on the negative diffusion loss, which allows DiffAIL to restrict the discriminator output strictly within $(0,1)$ and continue satisfying the optimization objective of minimizing the JS divergence. This is consistent with the logit value range after the sigmoid function in the original Gail and coincides with the monotonicity requirement.

By combining Eq.~\eqref{eq15} and Eq.~\eqref{eq16}, we can obtain our final optimization objective, which utilizes the diffusion loss as the part of discriminator in AIL:
\begin{align}\label{eq17}
    \min_{{\pi_\theta}}\max_{D_{\phi}} \mathbb{E}_{ \epsilon \sim \mathcal{N}, t \sim \mu} [ \notag  \mathbb{E}&_{x \sim {\pi_e} }[\log(\exp(-Diff_{\phi}(x,\epsilon,t)))] \\
      + \mathbb{E}_{x\sim \pi_\theta}[\log(&1- \exp(-Diff_{\phi}(x,\epsilon,t)))]].
\end{align}
In this adversarial framework, training the diffusion model loss is an adversarial process. $Diff_\phi(x,\epsilon,t)$ will minimize the diffusion error for $x$ from the expert $\pi_e$ and maximize for $x$ generated samples from the policy $\pi_{\theta}$. Therefore, $Diff_\phi(x,\epsilon,t)$ will learn to distinguish between expert demonstrations and policy yield samples. $\pi_\theta$ is a generator to generate state-action pairs similar to expert data, confusing the discriminator.

For outer minimization of Eq.\eqref{eq17}, it is usually to optimize $\pi_\theta$ by an arbitrary reinforcement learning algorithm with reward signal. So next, we use a diffusion-based discriminator to obtain our surrogate reward function like Gail~\cite{ho2016generative}:
\begin{equation}\label{eq18}
    R_\phi(x,\epsilon)=-\frac{1}{T}\sum_{t=1}^{T} \log(1-\exp{(-Diff_\phi(x,\epsilon,t)})) .
\end{equation}
During the training process for the diffusion loss, we perform uniform sampling from the interval $[1, T]$. When using diffusion loss as part of the surrogate reward function, we only need to compute the average loss of the diffusion model over $N$ timesteps instead of uniform sampling.
We give high rewards to the $x$ with low diffusion errors and vice versa. $\pi_\theta$ can conduct policy improvement from any forward reinforcement learning algorithms with the surrogate reward function. We summarize the overall process of the method in Algorithm~\ref{algorithm1}.

\begin{figure*}[htp]
\centering
\includegraphics[width=1.0\textwidth]{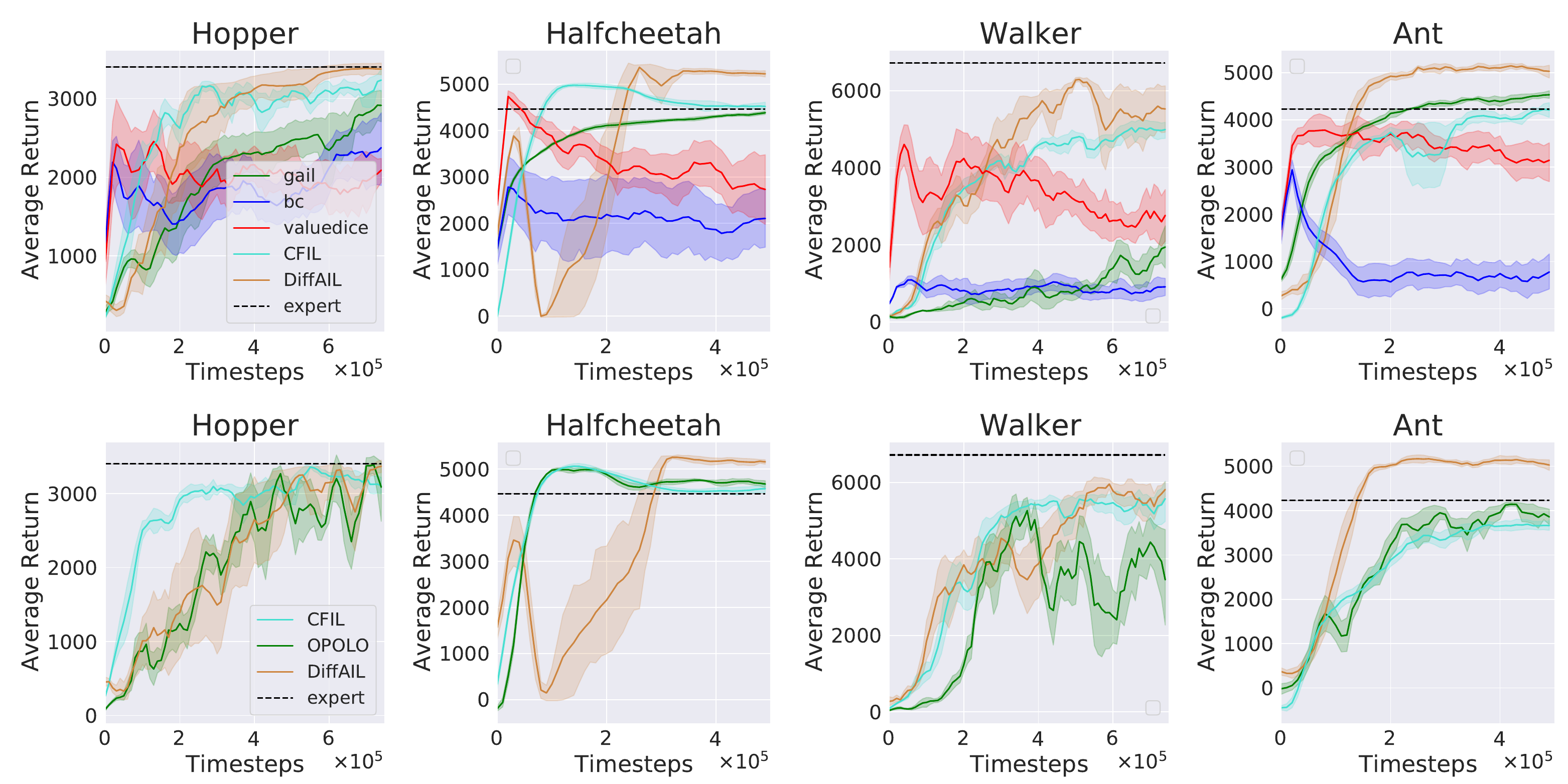}
\caption{\textbf{Top}: Learning curve for different sota imitation learning algorithms with 1 trajectory 5 five seeds in the standard state-action setting. \textbf{Bottom}: Learning curve for different sota imitation learning algorithms with one trajectory over five seeds in the state-only setting. The x-axis denotes timesteps, and the y-axis denotes the average return. The shadow areas represent the standard deviation.}
\label{1-trajectories}
\end{figure*}

\begin{algorithm}[ht]
\caption{Diffusion Adversarial Imitation Learning}
\label{algorithm1}
\textbf{Require}: initial policy network with parameter $\theta$, $Q$ network with parameter $\omega$ and diffusion discriminator network with parameter $\phi$, expert data $\mathcal{R}_e$, empty replay buffer $\mathcal{R}$, batch size $k$, total timesteps $N$, diffusion timesteps $T$, learning rate $\eta_\phi,\eta_\theta,\eta_\omega$ for parameters $\phi,\theta,\omega$ respectively.
\begin{algorithmic}[1]
\STATE let $n=0$.
\WHILE{$n < N $ }
\STATE Collect ($s,a,s'$) according to policy $\pi_\theta$ during interaction and add samples into $\mathcal{R}$.
\STATE Sample $\{(s_i^{\pi_\theta},a_i^{\pi_\theta})\}_{i=1}^k\sim \mathcal{R}$, $\{(s_i^{\pi_e},a_i^{\pi_e})\}_{i=1}^k\sim \mathcal{R}_e$.
\STATE Sample $\{(t_i)\}_{i=1}^k \sim \mu(1,T), \{(\epsilon_i)\}_{i=1}^k \sim \mathcal{N}(0,I)$ .
\STATE Compute the gradient of diffusion discriminator $\phi$:
\begin{equation*}
    \begin{aligned}
        ~ ~\nabla \mathcal{L}&=~\nabla_\phi\log (\exp(-Diff_\phi(s_i^{\pi_e},a_i^{\pi_e},\epsilon_i,t_i))) \\
        & + \nabla_\phi \log(1-\exp(-Diff_\phi(s_i^{\pi_\theta},a_i^{\pi_\theta},\epsilon_i,t_i))).
    \end{aligned}
\end{equation*}
\STATE Update diffusion discriminator $\phi$ by gradient ascent: \\ ~ ~ $\phi \leftarrow \phi + \eta_\phi\nabla \mathcal{L}$
\STATE Sample $\{(s_i^{\pi_\theta},a_i^{\pi_\theta})\}_{i=1}^k\sim \mathcal{R},\{(\epsilon_i)\}_{i=1}^k \sim \mathcal{N}(0,I)$.
\STATE Compute the surrogate reward $R_\phi(s_i^{\pi_\theta},a_i^{\pi_\theta},\epsilon_i)$ by Eq.~\eqref{eq18}.
\STATE Update SAC parameters $\pi_\theta$ and $Q_\omega$ by $\eta_\phi$ and $\eta_\omega$ with the surrogate reward.
\ENDWHILE
\end{algorithmic}
\end{algorithm}
\section{Related Work}

\subsection{Adversarial Imitation Learning}
AIL can be viewed as a unified perspective to minimize f-divergences, including Gail~\cite{ho2020denoising}, AIRL~\cite{fu2017learning}, FAIRL~\cite{ghasemipour2020divergence}, f-Gail~\cite{zhang2020f}, etc. These methods aim to minimize divergence by matching the occupancy measure of expert demonstrations and policy data. The policy improvement process and the discriminator optimization process are similar to the GAN~\cite{goodfellow2014generative}. However, these methods typically rely on on-policy sampling and often require millions or even tens of millions of interactions. To address this issue, DAC ~\cite{kostrikov2018discriminator} extends Gail to the off-policy algorithm, significantly improving the learning efficiency of AIL methods and reducing the necessary online interaction to hundreds of thousands of steps. AEAIL~\cite{zhang2022auto} proposes to use the reconstruction loss of Autoencoder as the reward function for AIL.

Additionally, a novel category of AIL methods, such as Valuedice~\cite{kostrikov2019imitation} and CFIL~\cite{freund2023coupled}, employ distribution correction estimation~\cite{nachum2019dualdice} to derive the objective function. They recover the log-density ratio by solving the optimal point of the Donsker-Varadhan representation of the KL divergence. Valuedice builds a fully off-policy optimization objective and outperforms BC in a strictly offline setting. However, recent work~\cite{li2022rethinking} proved that BC is optimal in the strictly offline setting, which raises doubts about the performance of Valuedice. CFIL argues that the log-density ratio cannot be accurately estimated without appropriate modelling tools and proposes using a coupled normalization flow to model the log-likelihood ratio explicitly. 
\subsection{Diffusion Model with Reinforcement Learning}
In RL, many methods leverage the powerful generative capabilities of diffusion models to generate state-action pairs or trajectories. Diffusion-BC~\cite{pearce2023imitating} addresses the limitation of insufficient expressiveness in traditional behavior cloning by modeling diffusion model as a policy. Diffuser~\cite{janner2022planning} uses an unconditional diffusion model to generate trajectories consisting of states and actions but requires an additional reward function to guide the reverse denoising process towards high-return trajectories. Diffuser Decision~\cite{ajay2022conditional} employs a conditional diffusion model with classifier-free guidance to model trajectories containing only states. However, since the modelling process does not involve actions, it is necessary to additionally learn an additional inverse dynamics model to select actions for any two adjacent states along the planned trajectory. BESO~\cite{reuss2023goal} applies the conditional diffusion model to goal-conditional imitation learning by directly modelling state and action as conditional distributions and using future states as conditions. Diffusion Q~\cite{wang2022diffusion} represents the diffusion model as a policy, explicitly regularizes it, and adds the maximum action value function term to the training objective to adjust the diffusion policy to select actions with the highest Q value.

Our method also utilizes the diffusion model, which differs from the abovementioned diffusion-based methods. Specifically, our approach remains within the AIL framework, unlike BESO, Diffuser Decision, and Diffuser, which directly employ the diffusion model as an expression policy rather than in standard reinforcement learning or imitation learning. Moreover, DiffAIL models an unconditional joint distribution that consists of state-action pairs instead of the conditional distribution.

\begin{figure*}[htp]
\centering
\includegraphics[width=1.0\textwidth]{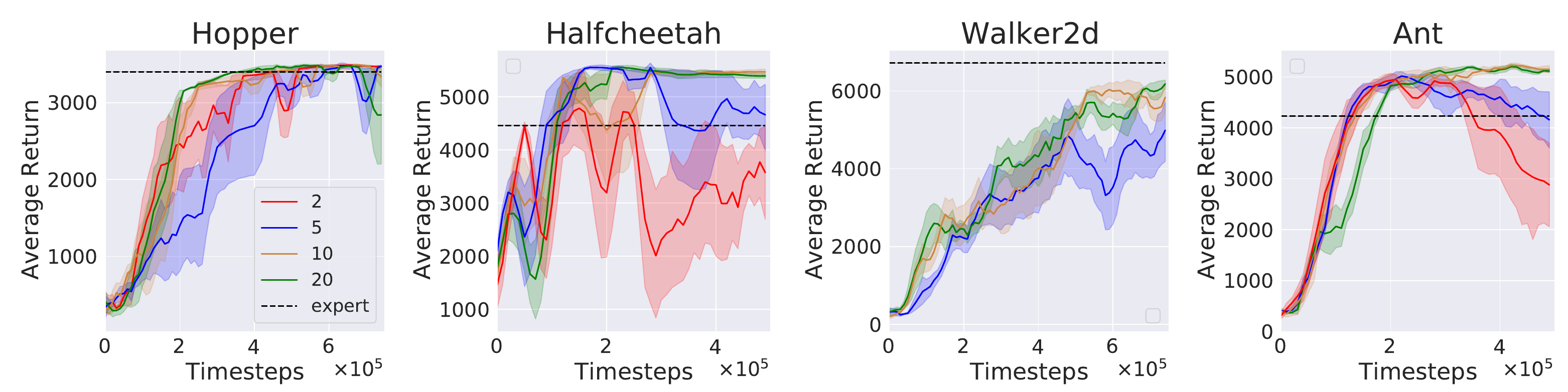} 
\caption{Ablation study on the diffusion timestamps of the diffusion discriminator using N grids [2, 5, 10, 20] with 4 trajectories over 5 seeds. Our findings suggest that a diffusion step of 10 yields good results for the diffusion discriminator. Note that at timestep 2, Walker2d did not show results for some random seeds due to training collapse.}
\label{diffusion_timesteps}
\end{figure*}

\section{Experiments}

In our experimental section, we explore the performance of the diffusion discriminator on various tasks. Our experiments focus on four key aspects: 
\begin{itemize} 
\item With the popular Mujoco environment, can our method achieve superior results with a small amount of demonstration data in state-action and state-only settings? 
\item On those expert-level unseen state-action pairs, can DiffAIL improve the generalization of successful identification for the discriminator compared to a naïve discriminator? 
\item Compared to naïve Gail, can DiffAIL provide a more linearly correlational surrogate reward? 
\item Since the number of diffusion steps significantly impacts model capability, will this phenomenon also occur in our methods?
\end{itemize}

\noindent\textbf{Benchmark \& Dataset.} In our experiments, we compare DiffAIL with the SOTA algorithm in the Mujoco environments. We choose four representative tasks, namely Hopper, HalfCheetah, Walker2d and Ant. We have $40$ trajectories on each task dataset, each with $1000$ steps. We randomly subsample $n=[1,4,16]$ trajectories from the pool of $40$ trajectories. Please refer to Appendix~A for detailed dataset descriptions.

\noindent\textbf{Baselines.} We benchmark our method against state-of-the-art imitation learning algorithms, including \textbf{(1)} Behavior cloning~\cite{pomerleau1991efficient}, the typical supervised learning approach; \textbf{(2)} Gail~\cite{ho2016generative}, the classical AIL method; \textbf{(3)} Valuedice~\cite{kostrikov2019imitation} which utilize distribution correction estimation; \textbf{(4)} CFIL~\cite{freund2023coupled}, which employs a coupled flow to evaluate the log-density ratio both used in state-action and state-only settings; \textbf{(5)} OPOLO~\cite{zhu2020off} adjusts policy updates by using a reverse dynamic model to accelerate learning for the state-only setting. 

\noindent\textbf{Implementation details.} Our denoising model uses a $3$-layer MLP with Mish activation function and $128$ hidden neurons in each layer rather than a standard U-Net architecture. The training uses the Adam optimizer\cite{kingma2014adam} with a learning rate of $0.00003$ and the gradient penalty\cite{gulrajani2017improved} for stability. Besides bc and Valuedice, other methods all use SAC\cite{haarnoja2018soft} for forward reinforcement learning. For more implementation details, please refer to Appendix~B. 

For a fair comparison,  we implemented an off-policy version of Gail to improve its sample efficiency, but it did not involve special processing of the bias of the absorbed state reward function in DAC~\cite{kostrikov2018discriminator}. All algorithms are evaluated under the same experimental setting. Experiments use five random seeds $(0,1,2,3,4)$, run for an equal number of steps per task, with $10$ thousand steps per epoch. Model performance is evaluated every $10$ thousand steps by averaging rewards over ten episodes. We then smooth the results using a sliding window of $4$ before plotting means and standard deviations across the $5$ seeds.

\subsection{Comparison to SOTA methods}

We conduct experiments on multiple tasks from the popular MuJoCo benchmark, using $1$, $4$, and $16$ expert trajectories to evaluate performance across varying trajectory numbers. For each algorithm, we assess the real average episodic return the agent achieves when interacting in the environment.

With only $1$ expert trajectory, the top of Figure~\ref{1-trajectories} shows our method outperforms other imitation learning baselines. Our approach achieves near expert-level asymptotic performance on all tasks and even significantly surpasses the expert on HalfCheetah and Ant. This expert-surpassing result is not performed by other methods. Especially table~\ref{table_1_traj} clearly shows that our method outperforms all other methods significantly on best performance during training.

Subsequently, we conducted our experiment under a state-only setting and found that the advantages brought by the powerful discriminator are very significant, not just in modeling state-action distribution. Under the state-only setting, DiffAIL models the state and the next state distribution. Importantly, we don't need to modify any hyperparameters, and then we directly validate the performance of our method. We compare with OPPLO and CFIL. From the bottom of Figure~\ref{1-trajectories}, it can be seen that our method can reach expert-level performance with 1 trajectory and outperform other methods, especially on HalfCheetah and Ant.

All of these significant advantages come from the diffusion discriminator being able to accurately capture the data distribution of experts, whether it is state action pairs or state next state pairs.

While our method achieves expert-level performance on multiple tasks, it is less time-efficient than Valuedice's optimized implementation. We acknowledge Valuedice's impressive efficiency gains, though algorithmic speed is not our primary focus. However, in one trajectory situation, we find it can't maintain optimality in later training stages stably. Still, our approach performs similarly or even more efficiently compared to other methods, except for Valuedice.

The experimental results of training additional $4$ and $16$ trajectories can be seen in Appendix~E.
\subsection{Discriminate on unused expert trajectories}

\begin{table}[t]
\centering
\begin{tabular}{c|c|c}
\toprule
Env & Gail & DiffAIL \\
\midrule
Hopper & 83.3\% ± 1.2\% & \textbf{92.8\% ± 2.5\%}  \\
HalfCheetah & 83.8\% ± 0.9\% & \textbf{94.5\% ± 0.6\%}  \\
Walker2d & 90.5\% ± 0.8\% & \textbf{96.1\% ± 0.6\%}  \\ 
Ant &  81.7\% ± 1.6\% & \textbf{98.8\% ± 0.1\%} \\ 
\bottomrule
\end{tabular}
\caption{A comparison of the ability to distinguish expert demonstrations out of data for Gail and DiffAIL with four trajectories over five seeds. It suggests that DiffAIL has a significant improvement over Gail's discriminator. }
\label{table_percent}
\end{table}

We conduct an additional experiment to verify that our method improves the generalization of unseen expert-level state-action pairs for discrimination. We take the discriminator 
which has completed training on just $4$ trajectories, and test it on $36$ held-out expert trajectories from the remaining dataset. If the discriminator can still successfully identify these new trajectories as expert-like, it indicates improved generalization. 

However, selecting an evaluation metric is challenging. Theoretically, the discriminator will eventually converge to $0.5$\cite{goodfellow2014generative}, where it cannot distinguish between expert demonstrations and policy data. But in practice, training ends before the discriminator and generator reach the ideal optimum, and the discriminator often overfits the expert demonstrations. Therefore, we adopted a simple evaluation criterion: if the discriminator output is greater than or equal to $ 0.5 $, the expert data is considered successfully distinguished; otherwise, the discrimination fails.

As shown in Table~\ref{table_percent}, it represents the percentage of state-action pairs successfully distinguished by the discriminator from $36$ trajectories, i.e., $36000$ state-action pairs. From the benchmark of HalfCheetah and Ant, DiffAIL can significantly improve the discrimination generalization of unseen expert demonstrations. In other tasks, DiffAIL also exhibits stronger discriminative abilities than Gail, which sufficiently exhibits the powerful advantage of the diffusion discriminator in capturing the expert distribution. Notably, all trajectories of the same task are yielded by interactions between the same trained policy. Therefore, substantial similarities may exist across state-action pairs from different trajectories, which also explains why Gail can achieve decent discrimination as well.

\begin{figure}[htp]
    \begin{minipage}{0.48\textwidth}
    \centering
    \includegraphics[width=\textwidth]{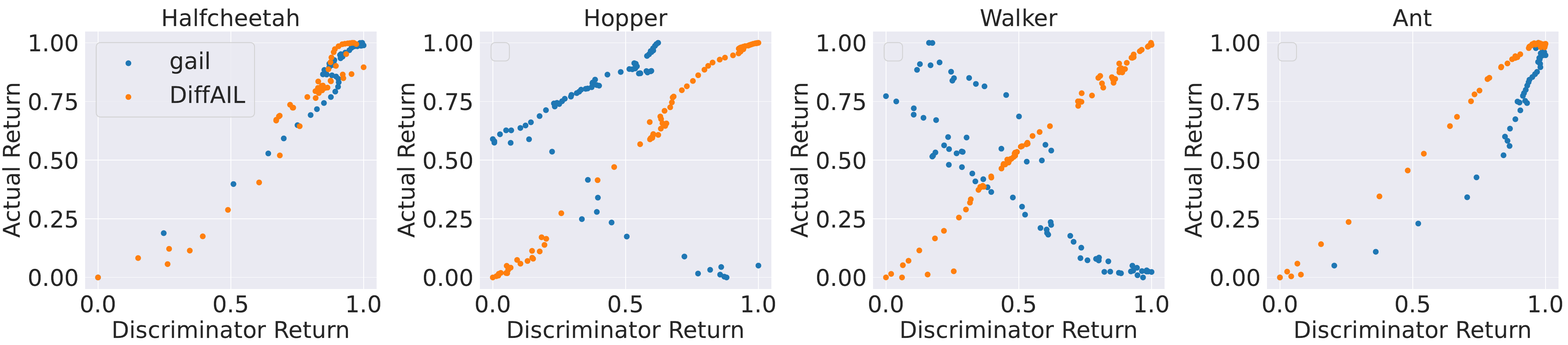}
    
    \end{minipage}
    \label{Correlatioin}
    \caption{Correlation between average discriminator return and average actual return with Gail and DiffAIL in 3 seeds. The returns have been normalized for display convenience.}
\end{figure}

\subsection{Return correlations}\label{Return correlation}
As Eq.~\eqref{eq3} shows, a better discriminator can provide better surrogate rewards to guide forward reinforcement learning. We verify this by experimentally analyzing the Pearson coefficient(PC) between the surrogate discriminator return and the actual return. We compare ours with the vanilla discriminator in GAIL, as shown in Figure~3: The PC for DiffAIL are $[ 97.1\%, 98.8\%, 99.1\%, 99.6\%]$, while for GAIL are $[ 97.0\%, 54.1\%, -89.0\%, 92.1\%]$. This indicates that our diffusion surrogate reward function is more linearly correlated with actual rewards than GAIL, which can better guide policy learning.

\subsection{Ablation study}

We conduct ablation studies to investigate how various diffusion steps impact the distribution-matching ability of the diffusion discriminator on state-action pairs. 

Figure~\ref{diffusion_timesteps} shows that we set the diffusion steps to $t=[2,5,10,20]$, respectively. The results show that increasing diffusion steps leads to more stable performance and faster convergence across all tasks, aligning with results in diffusion models for image generation. However, we also observe that more diffusion steps increase training time, as exhibited in Appendix (Table~4). Therefore, we finally select $t=10$ as the diffusion step to balance discriminator quality and training cost.
\section{Conclusion}

 In this work, we propose DiffAIL, which utilizes diffusion models for AIL. Unlike prior methods that directly apply diffusion models as an expressive policy, our approach leverages the diffusion model to enhance the distribution matching ability of the discriminator, which enables the discriminator to improve generalization for expert-level state-action pairs. In both the state-action setting and state-only setting, We model the joint distribution of state-action pairs and state and next state pairs, respectively, with an unconditional diffusion model. Surprisingly, in both settings with a single expert trajectory, the result shows our method achieves SOTA performance across all tasks. Additionally, we discuss the limitations of this work and some promising future directions in Appendix~D.

\section{Acknowledgments}

This work was supported by the NSFC (Nos. 62206159, 62176139), the Natural Science Foundation of Shandong Province (Nos. ZR2022QF117, ZR2021ZD15), the Fundamental Research Funds of Shandong University, the Fundamental Research Funds for the Central Universities. G. Wu was sponsored by the TaiShan Scholars Program.

\bibliography{aaai24}

\begin{thebibliography}{42}
\providecommand{\natexlab}[1]{#1}

\bibitem[{Ajay et~al.(2022)Ajay, Du, Gupta, Tenenbaum, Jaakkola, and Agrawal}]{ajay2022conditional}
Ajay, A.; Du, Y.; Gupta, A.; Tenenbaum, J.; Jaakkola, T.; and Agrawal, P. 2022.
\newblock Is conditional generative modeling all you need for decision-making?
\newblock \emph{arXiv preprint arXiv:2211.15657}.

\bibitem[{Bao et~al.(2022)Bao, Li, Zhu, and Zhang}]{bao2022analytic}
Bao, F.; Li, C.; Zhu, J.; and Zhang, B. 2022.
\newblock Analytic-dpm: an analytic estimate of the optimal reverse variance in diffusion probabilistic models.
\newblock \emph{arXiv preprint arXiv:2201.06503}.

\bibitem[{Freund, Sarafian, and Kraus(2023)}]{freund2023coupled}
Freund, G.; Sarafian, E.; and Kraus, S. 2023.
\newblock A Coupled Flow Approach to Imitation Learning.
\newblock \emph{arXiv preprint arXiv:2305.00303}.

\bibitem[{Fu, Luo, and Levine(2017)}]{fu2017learning}
Fu, J.; Luo, K.; and Levine, S. 2017.
\newblock Learning robust rewards with adversarial inverse reinforcement learning.
\newblock \emph{arXiv preprint arXiv:1710.11248}.

\bibitem[{Fujimoto, Hoof, and Meger(2018)}]{fujimoto2018addressing}
Fujimoto, S.; Hoof, H.; and Meger, D. 2018.
\newblock Addressing function approximation error in actor-critic methods.
\newblock In \emph{International conference on machine learning}, 1587--1596. PMLR.

\bibitem[{Garg et~al.(2021)Garg, Chakraborty, Cundy, Song, and Ermon}]{garg2021iq}
Garg, D.; Chakraborty, S.; Cundy, C.; Song, J.; and Ermon, S. 2021.
\newblock Iq-learn: Inverse soft-q learning for imitation.
\newblock \emph{Advances in Neural Information Processing Systems}, 34: 4028--4039.

\bibitem[{Ghasemipour, Zemel, and Gu(2020)}]{ghasemipour2020divergence}
Ghasemipour, S. K.~S.; Zemel, R.; and Gu, S. 2020.
\newblock A divergence minimization perspective on imitation learning methods.
\newblock In \emph{Conference on Robot Learning}, 1259--1277. PMLR.

\bibitem[{Goodfellow et~al.(2014)Goodfellow, Pouget-Abadie, Mirza, Xu, Warde-Farley, Ozair, Courville, and Bengio}]{goodfellow2014generative}
Goodfellow, I.; Pouget-Abadie, J.; Mirza, M.; Xu, B.; Warde-Farley, D.; Ozair, S.; Courville, A.; and Bengio, Y. 2014.
\newblock Generative adversarial nets.
\newblock \emph{Advances in neural information processing systems}, 27.

\bibitem[{Gulrajani et~al.(2017)Gulrajani, Ahmed, Arjovsky, Dumoulin, and Courville}]{gulrajani2017improved}
Gulrajani, I.; Ahmed, F.; Arjovsky, M.; Dumoulin, V.; and Courville, A.~C. 2017.
\newblock Improved training of wasserstein gans.
\newblock \emph{Advances in neural information processing systems}, 30.

\bibitem[{Haarnoja et~al.(2018)Haarnoja, Zhou, Abbeel, and Levine}]{haarnoja2018soft}
Haarnoja, T.; Zhou, A.; Abbeel, P.; and Levine, S. 2018.
\newblock Soft actor-critic: Off-policy maximum entropy deep reinforcement learning with a stochastic actor.
\newblock In \emph{International conference on machine learning}, 1861--1870. PMLR.

\bibitem[{Ho and Ermon(2016)}]{ho2016generative}
Ho, J.; and Ermon, S. 2016.
\newblock Generative adversarial imitation learning.
\newblock \emph{Advances in neural information processing systems}, 29.

\bibitem[{Ho, Jain, and Abbeel(2020)}]{ho2020denoising}
Ho, J.; Jain, A.; and Abbeel, P. 2020.
\newblock Denoising diffusion probabilistic models.
\newblock \emph{Advances in Neural Information Processing Systems}, 33: 6840--6851.

\bibitem[{Janner et~al.(2022)Janner, Du, Tenenbaum, and Levine}]{janner2022planning}
Janner, M.; Du, Y.; Tenenbaum, J.~B.; and Levine, S. 2022.
\newblock Planning with diffusion for flexible behavior synthesis.
\newblock \emph{arXiv preprint arXiv:2205.09991}.

\bibitem[{Kingma and Ba(2014)}]{kingma2014adam}
Kingma, D.~P.; and Ba, J. 2014.
\newblock Adam: A method for stochastic optimization.
\newblock \emph{arXiv preprint arXiv:1412.6980}.

\bibitem[{Kingma and Welling(2013)}]{kingma2013auto}
Kingma, D.~P.; and Welling, M. 2013.
\newblock Auto-encoding variational bayes.
\newblock \emph{arXiv preprint arXiv:1312.6114}.

\bibitem[{Kostrikov et~al.(2018)Kostrikov, Agrawal, Dwibedi, Levine, and Tompson}]{kostrikov2018discriminator}
Kostrikov, I.; Agrawal, K.~K.; Dwibedi, D.; Levine, S.; and Tompson, J. 2018.
\newblock Discriminator-actor-critic: Addressing sample inefficiency and reward bias in adversarial imitation learning.
\newblock \emph{arXiv preprint arXiv:1809.02925}.

\bibitem[{Kostrikov, Nachum, and Tompson(2019)}]{kostrikov2019imitation}
Kostrikov, I.; Nachum, O.; and Tompson, J. 2019.
\newblock Imitation learning via off-policy distribution matching.
\newblock \emph{arXiv preprint arXiv:1912.05032}.

\bibitem[{Li et~al.(2022)Li, Xu, Yu, and Luo}]{li2022rethinking}
Li, Z.; Xu, T.; Yu, Y.; and Luo, Z.-Q. 2022.
\newblock Rethinking ValueDice: Does It Really Improve Performance?
\newblock \emph{arXiv preprint arXiv:2202.02468}.

\bibitem[{Liu et~al.(2022)Liu, Ren, Lin, and Zhao}]{liu2022pseudo}
Liu, L.; Ren, Y.; Lin, Z.; and Zhao, Z. 2022.
\newblock Pseudo numerical methods for diffusion models on manifolds.
\newblock \emph{arXiv preprint arXiv:2202.09778}.

\bibitem[{Liu, Gong, and Liu(2022)}]{liu2022flow}
Liu, X.; Gong, C.; and Liu, Q. 2022.
\newblock Flow straight and fast: Learning to generate and transfer data with rectified flow.
\newblock \emph{arXiv preprint arXiv:2209.03003}.

\bibitem[{Liu et~al.(2023)Liu, Xu, Zhang, Liu, Jiang, Chen, Zhang, and Yu}]{liu2023guide}
Liu, X.-H.; Xu, F.; Zhang, X.; Liu, T.; Jiang, S.; Chen, R.; Zhang, Z.; and Yu, Y. 2023.
\newblock How To Guide Your Learner: Imitation Learning with Active Adaptive Expert Involvement.
\newblock \emph{arXiv preprint arXiv:2303.02073}.

\bibitem[{Lu et~al.(2022{\natexlab{a}})Lu, Zhou, Bao, Chen, Li, and Zhu}]{lu2022dpm}
Lu, C.; Zhou, Y.; Bao, F.; Chen, J.; Li, C.; and Zhu, J. 2022{\natexlab{a}}.
\newblock Dpm-solver: A fast ode solver for diffusion probabilistic model sampling in around 10 steps.
\newblock \emph{Advances in Neural Information Processing Systems}, 35: 5775--5787.

\bibitem[{Lu et~al.(2022{\natexlab{b}})Lu, Zhou, Bao, Chen, Li, and Zhu}]{lu2022dpm++}
Lu, C.; Zhou, Y.; Bao, F.; Chen, J.; Li, C.; and Zhu, J. 2022{\natexlab{b}}.
\newblock Dpm-solver++: Fast solver for guided sampling of diffusion probabilistic models.
\newblock \emph{arXiv preprint arXiv:2211.01095}.

\bibitem[{Mnih et~al.(2013)Mnih, Kavukcuoglu, Silver, Graves, Antonoglou, Wierstra, and Riedmiller}]{mnih2013playing}
Mnih, V.; Kavukcuoglu, K.; Silver, D.; Graves, A.; Antonoglou, I.; Wierstra, D.; and Riedmiller, M. 2013.
\newblock Playing atari with deep reinforcement learning.
\newblock \emph{arXiv preprint arXiv:1312.5602}.

\bibitem[{Mnih et~al.(2015)Mnih, Kavukcuoglu, Silver, Rusu, Veness, Bellemare, Graves, Riedmiller, Fidjeland, Ostrovski et~al.}]{mnih2015human}
Mnih, V.; Kavukcuoglu, K.; Silver, D.; Rusu, A.~A.; Veness, J.; Bellemare, M.~G.; Graves, A.; Riedmiller, M.; Fidjeland, A.~K.; Ostrovski, G.; et~al. 2015.
\newblock Human-level control through deep reinforcement learning.
\newblock \emph{nature}, 518(7540): 529--533.

\bibitem[{Nachum et~al.(2019)Nachum, Chow, Dai, and Li}]{nachum2019dualdice}
Nachum, O.; Chow, Y.; Dai, B.; and Li, L. 2019.
\newblock Dualdice: Behavior-agnostic estimation of discounted stationary distribution corrections.
\newblock \emph{Advances in Neural Information Processing Systems}, 32.

\bibitem[{Nowozin, Cseke, and Tomioka(2016)}]{nowozin2016f}
Nowozin, S.; Cseke, B.; and Tomioka, R. 2016.
\newblock f-gan: Training generative neural samplers using variational divergence minimization.
\newblock \emph{Advances in neural information processing systems}, 29.

\bibitem[{Pearce et~al.(2023)Pearce, Rashid, Kanervisto, Bignell, Sun, Georgescu, Macua, Tan, Momennejad, Hofmann et~al.}]{pearce2023imitating}
Pearce, T.; Rashid, T.; Kanervisto, A.; Bignell, D.; Sun, M.; Georgescu, R.; Macua, S.~V.; Tan, S.~Z.; Momennejad, I.; Hofmann, K.; et~al. 2023.
\newblock Imitating human behaviour with diffusion models.
\newblock \emph{arXiv preprint arXiv:2301.10677}.

\bibitem[{Pomerleau(1991)}]{pomerleau1991efficient}
Pomerleau, D.~A. 1991.
\newblock Efficient training of artificial neural networks for autonomous navigation.
\newblock \emph{Neural computation}, 3(1): 88--97.

\bibitem[{Reuss et~al.(2023)Reuss, Li, Jia, and Lioutikov}]{reuss2023goal}
Reuss, M.; Li, M.; Jia, X.; and Lioutikov, R. 2023.
\newblock Goal-conditioned imitation learning using score-based diffusion policies.
\newblock \emph{arXiv preprint arXiv:2304.02532}.

\bibitem[{Ross, Gordon, and Bagnell(2011)}]{ross2011reduction}
Ross, S.; Gordon, G.; and Bagnell, D. 2011.
\newblock A reduction of imitation learning and structured prediction to no-regret online learning.
\newblock In \emph{Proceedings of the fourteenth international conference on artificial intelligence and statistics}, 627--635. JMLR Workshop and Conference Proceedings.

\bibitem[{Schulman et~al.(2015)Schulman, Levine, Abbeel, Jordan, and Moritz}]{schulman2015trust}
Schulman, J.; Levine, S.; Abbeel, P.; Jordan, M.; and Moritz, P. 2015.
\newblock Trust region policy optimization.
\newblock In \emph{International conference on machine learning}, 1889--1897. PMLR.

\bibitem[{Schulman et~al.(2017)Schulman, Wolski, Dhariwal, Radford, and Klimov}]{schulman2017proximal}
Schulman, J.; Wolski, F.; Dhariwal, P.; Radford, A.; and Klimov, O. 2017.
\newblock Proximal policy optimization algorithms.
\newblock \emph{arXiv preprint arXiv:1707.06347}.

\bibitem[{Silver et~al.(2016)Silver, Huang, Maddison, Guez, Sifre, Van Den~Driessche, Schrittwieser, Antonoglou, Panneershelvam, Lanctot et~al.}]{silver2016mastering}
Silver, D.; Huang, A.; Maddison, C.~J.; Guez, A.; Sifre, L.; Van Den~Driessche, G.; Schrittwieser, J.; Antonoglou, I.; Panneershelvam, V.; Lanctot, M.; et~al. 2016.
\newblock Mastering the game of Go with deep neural networks and tree search.
\newblock \emph{nature}, 529(7587): 484--489.

\bibitem[{Song, Meng, and Ermon(2020)}]{song2020denoising}
Song, J.; Meng, C.; and Ermon, S. 2020.
\newblock Denoising diffusion implicit models.
\newblock \emph{arXiv preprint arXiv:2010.02502}.

\bibitem[{Song et~al.(2023)Song, Dhariwal, Chen, and Sutskever}]{song2023consistency}
Song, Y.; Dhariwal, P.; Chen, M.; and Sutskever, I. 2023.
\newblock Consistency Models.
\newblock \emph{arXiv preprint arXiv:2303.01469}.

\bibitem[{Todorov, Erez, and Tassa(2012)}]{todorov2012mujoco}
Todorov, E.; Erez, T.; and Tassa, Y. 2012.
\newblock Mujoco: A physics engine for model-based control.
\newblock In \emph{2012 IEEE/RSJ international conference on intelligent robots and systems}, 5026--5033. IEEE.

\bibitem[{Wang, Hunt, and Zhou(2022)}]{wang2022diffusion}
Wang, Z.; Hunt, J.~J.; and Zhou, M. 2022.
\newblock Diffusion policies as an expressive policy class for offline reinforcement learning.
\newblock \emph{arXiv preprint arXiv:2208.06193}.

\bibitem[{Zhang et~al.(2022)Zhang, Zhao, Zhang, and Gao}]{zhang2022auto}
Zhang, K.; Zhao, R.; Zhang, Z.; and Gao, Y. 2022.
\newblock Auto-Encoding Adversarial Imitation Learning.

\bibitem[{Zhang et~al.(2020)Zhang, Li, Zhang, and Zhang}]{zhang2020f}
Zhang, X.; Li, Y.; Zhang, Z.; and Zhang, Z.-L. 2020.
\newblock f-gail: Learning f-divergence for generative adversarial imitation learning.
\newblock \emph{Advances in neural information processing systems}, 33: 12805--12815.

\bibitem[{Zheng et~al.(2023)Zheng, Lu, Chen, and Zhu}]{zheng2023dpm}
Zheng, K.; Lu, C.; Chen, J.; and Zhu, J. 2023.
\newblock DPM-Solver-v3: Improved Diffusion ODE Solver with Empirical Model Statistics.
\newblock \emph{arXiv preprint arXiv:2310.13268}.

\bibitem[{Zhu et~al.(2020)Zhu, Lin, Dai, and Zhou}]{zhu2020off}
Zhu, Z.; Lin, K.; Dai, B.; and Zhou, J. 2020.
\newblock Off-policy imitation learning from observations.
\newblock \emph{Advances in Neural Information Processing Systems}, 33: 12402--12413.

\end{thebibliography}


\clearpage
\section{Appendices}
\begin{table*}[htp]
\setcounter{table}{2}
\centering
\begin{tabular}{c|c|c|c|c}
\toprule
Tasks & HalfCheetah & Hopper & Walker2d & Ant  \\
\midrule
Optimizer & Adam & Adam & Adam & Adam\\
Batch Size & $256$ & $256$  & $256$ & $256$  \\ 
diffusion hidden layer & $128$ & $128$  & $128$ & $128$  \\ 
diffusion timesteps & $10$ & $10$ & $10$ & $10$ \\ 
noise schedule & linear & linear & linear & linear \\ 
Diffusion learning rate & $3 \times 10^{-4}$ & $3 \times 10^{-4}$  & $3 \times 10^{-4}$ & $3 \times 10^{-4}$ \\ 
use gradient penalty & $\textbf{True}$ & $\textbf{False}$ & $\textbf{True}$ & $\textbf{True}$ \\ 
gradient penalty weight & $\textbf{0.1}$ & $\textbf{NAN}$ & $\textbf{0.001}$ & $\textbf{0.1}$ \\ 
Discount factor $\gamma$ &$0.99$ & $0.99$  & $0.99$ & $0.99$ \\
$Q$ learning rate &  $3 \times 10^{-3}$ & $3 \times 10^{-3}$  & $3 \times 10^{-3}$ & $3 \times 10^{-3}$\\ 
$\pi$ learning rate & $3 \times 10^{-3}$ & $3 \times 10^{-3}$  & $3 \times 10^{-3}$ & $3 \times 10^{-3}$ \\ 
\bottomrule
\end{tabular}
\caption{Optimal hyperparameters of DiffAIL in both state-action and state-only settings. DiffAIL parameters change only on the gradient penalty term, indicating that it is very applicable. }
\label{table_optimal_hyperparameters}
\end{table*}

\begin{table*}[htp]
\centering
\begin{tabular}{c|c|c|c|c|c}
\toprule
diffusion steps & Gail & DiffAIL$(2)$ & DiffAIL$(5)$ & DiffAIL$(10)$ & DiffAIL$(20)$ \\
\midrule
HalfCheetah $(timesteps=50)$ & 3.1h & 5.5h & 5.9h & 6.2h & 8.0h  \\
Ant $(timesteps=50)$ &  3.2h & 5.7h & 6.1 & 6.4 & 8.2h\\ 
Hopper $(timesteps=75)$ &4.5h & 6.5h & 6.9h & 7.5h & 9.3h \\
Walker2d $(timesteps=75)$ &4.8h & 8.4h &  8.9h & 9.5h & 11.4h \\ 
\bottomrule
\end{tabular}
\caption{A comparison of running time for different diffusion steps 
 with DiffAIL on $4$ Mujoco Tasks, with a unit equal to $1$ hour. The first line denotes the algorithms, where the numbers in parentheses represent the diffusion steps. The first column represents the Mujoco task, where the number in parentheses represents the training steps in units of $10,000$.}
\label{table_time}
\end{table*}

\begin{figure*}[htpb]
\setcounter{figure}{3}
\centering
\includegraphics[width=1.0\textwidth]{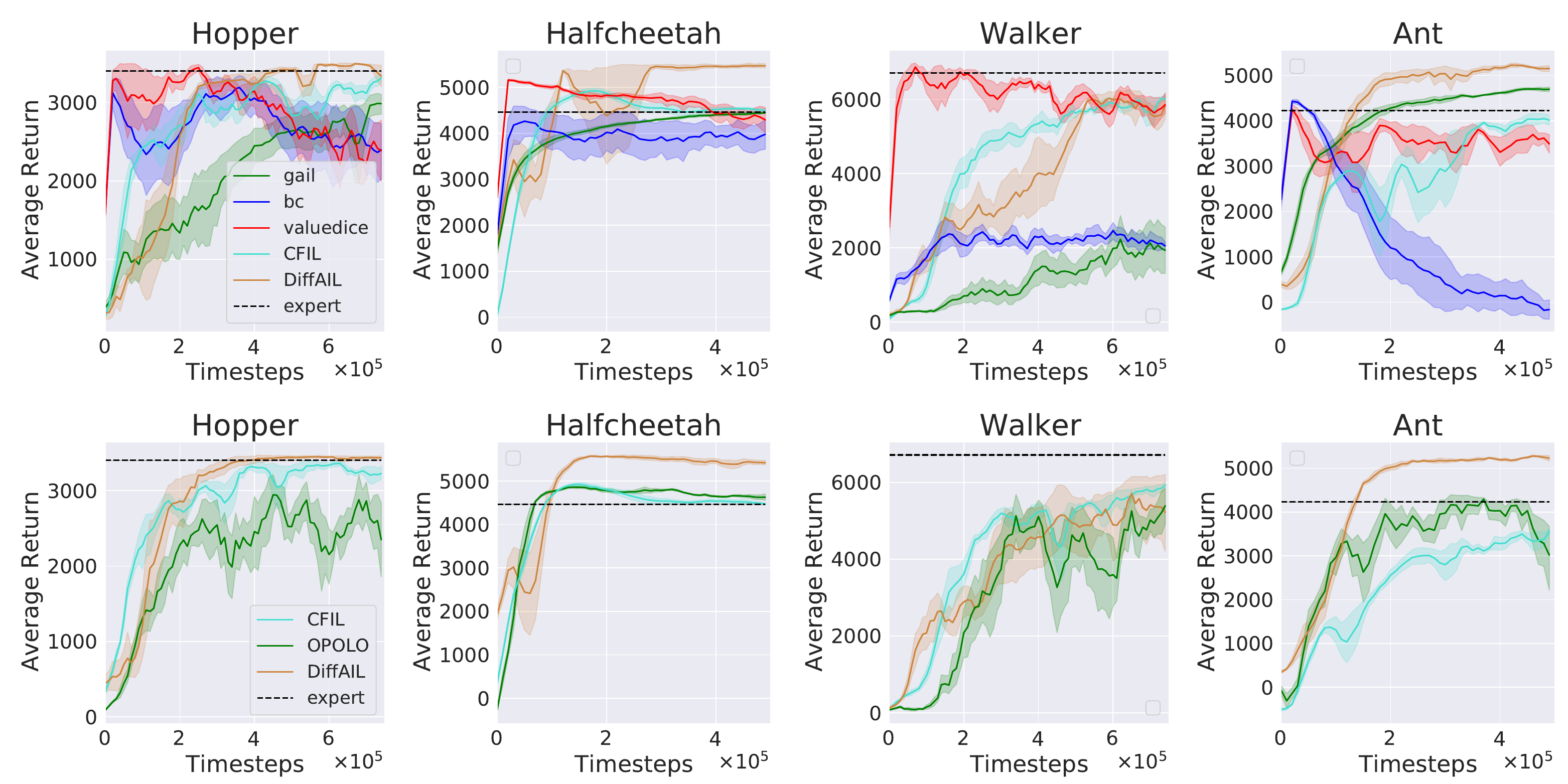} 
\caption{\textbf{Top}: Learning curve for different sota imitation learning algorithms with $4$ trajectories over $5$ seeds in the standard state-action setting. \textbf{Bottom}: Learning curve for different sota imitation learning algorithms with $4$ trajectories over $5$ seeds in the state-only setting. The x-axis denotes timesteps, and the y-axis denotes the average return. The shadow areas represent the standard deviation.}
\label{4-trajectories}
\end{figure*}

\begin{figure*}[htpb]
\centering
\includegraphics[width=1.0\textwidth]{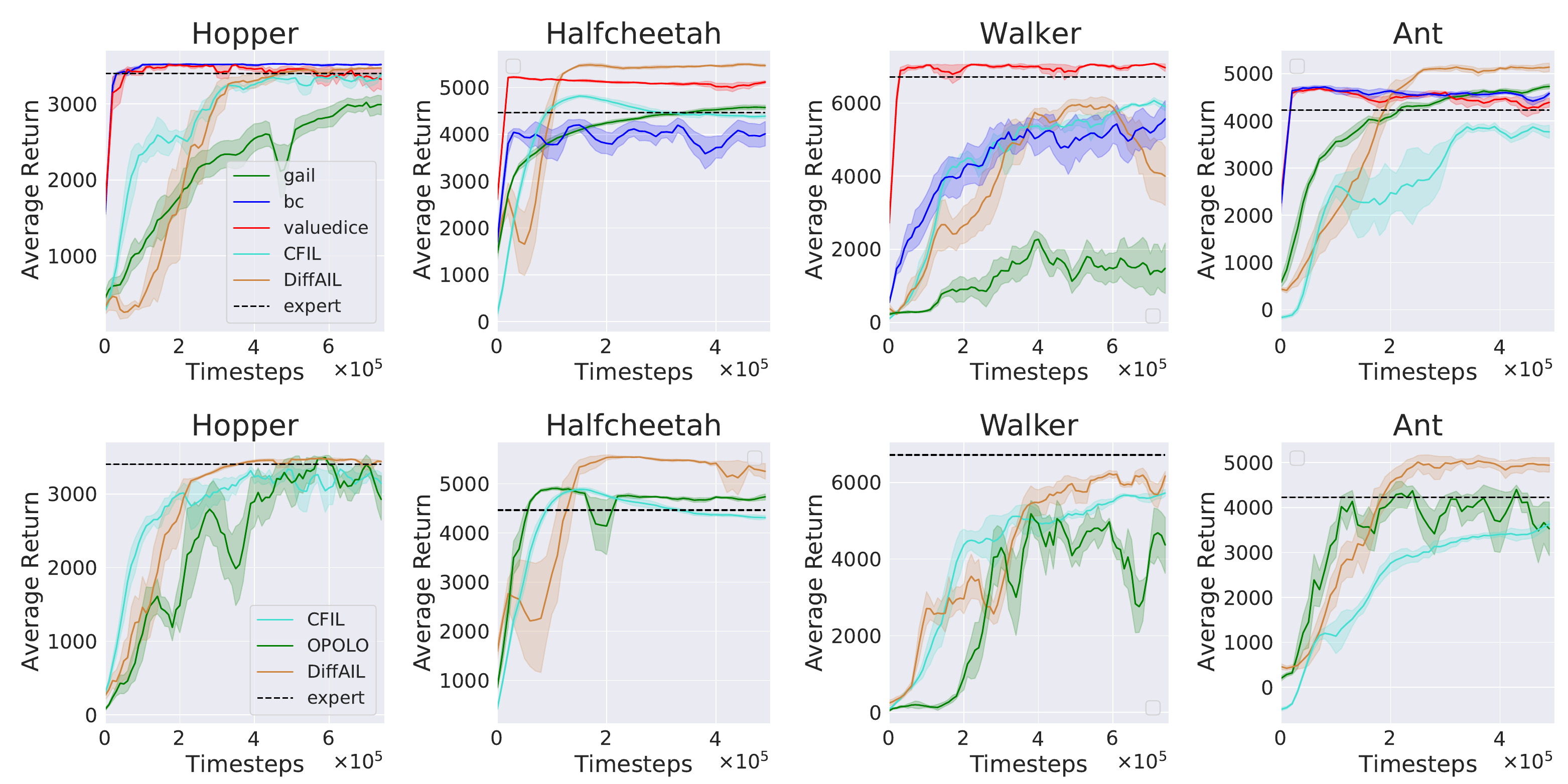}
\caption{\textbf{Top}: Learning curve for different sota imitation learning algorithms with $16$ trajectories over $5$ seeds in the standard state-action setting. \textbf{Bottom}: Learning curve for different sota imitation learning algorithms with $16$ trajectories over $5$ seeds in the state-only setting. The x-axis denotes timesteps, and the y-axis denotes the average return. The shadow areas represent the standard deviation.}
\label{16-trajectories}
\end{figure*}

\subsection{A \ \ \ Dataset discription}\label{Dataset-discription}
We utilize benchmark datasets from two sources: The Halfcheetah, Walker2d and Ant datasets are selected from the open-source valuedice\cite{kostrikov2019imitation} repository. For the Hopper, we use the SAC algorithm to interact with the environment for $1$ million steps to obtain an expert policy. We then collect $40$ expert trajectories by executing this policy in the environment. Since adversarial imitation learning requires a small amount of expert data, we subsample n trajectories from $40$ trajectories. We test $n=[1,4,16]$ to evaluate performance with varying trajectory set sizes.

\subsection{B \ \ \ Implementation details}\label{Implementation-details}
We build Gail\cite{ho2016generative} and DiffAIL on top of the implementation from the FAIRL\cite{ghasemipour2020divergence}\footnote{https://github.com/KamyarGh/rl\_swiss}. For the diffusion model in DiffAIL, we modify the code of Diffusion Q\cite{wang2022diffusion}\footnote{https://github.com/zhendong-wang/diffusion-policies-for-offline-rl} to create an unconditional variant. Our denoising model uses a $3$-layer MLP with Mish activations and $128$ hidden units rather than a U-Net architecture. We embed the timestep with a linear function. The training uses the Adam optimizer\cite{kingma2014adam} with a learning rate of $0.00003$ and the gradient penalty\cite{gulrajani2017improved} for stability. Besides bc and Valuedice, other methods use sac for forward reinforcement learning. The specific optimal parameters of DiffAIL can be seen in Table~\ref{table_optimal_hyperparameters}. For the baselines, we leverage open-source implementations of BC, Valuedice\footnote{https://github.com/google-research/google-research/tree/master/value\_dice}, and CFIL\footnote{https://github.com/gfreund123/cfil}, OPOLO\footnote{https://github.com/illidanlab/opolo-code}, which allow rapid benchmarking against prior state-of-the-art methods. We implemented all our experiments on a server with $8$ Tesla P100 graphics cards.

\subsection{C  \ \ \ Optimal hyperparameters}\label{Optimal-hyperparameters}

For the DiffAIL experiments, we summarize the optimal hyperparameters in Table~\ref{table_optimal_hyperparameters}. We mainly considered two hyperparameters - the learning rate of the diffusion model and the gradient penalty for the discriminator. The learning rate of the diffusion model was determined via grid search over $[0.0003, 0.00003,0.000003]$, and the gradient penalty was determined via grid search over $[0.001, 0.01, 0.1]$. For forward reinforcement learning, we employ the SAC algorithm with fixed parameters across all tasks.

\subsection{D \ \ \ Limitaion and future work}\label{limitation and future wrok}
In this work, we introduced diffusion models into the adversarial imitation learning framework and proposed DiffAIL. Experiments show that this approach could significantly enhance adversarial imitation learning performance, owing to the powerful distribution-capturing capability of the diffusion discriminator. Although sampling inference is unnecessary during this adversarial training, computing the noise mse loss $Diff_\phi$ still requires multi-step noise sampling, which undoubtedly increases the computational cost of DiffAIL. However, our ablation studies on diffusion steps showed that more steps lead to better discriminator training, posing a trade-off between the two. Meanwhile, we also noticed that much research has focused on accelerating sampling\cite{song2020denoising,bao2022analytic,lu2022dpm,liu2022flow,liu2022pseudo,song2023consistency} in diffusion models - these methods can maintain generation quality while drastically reducing sampling steps, even enabling one-step generation\cite{liu2022flow,song2023consistency}. Therefore, future Diffusion AIL could leverage advanced samplers to accelerate training and reduce computational costs.

Additionally, we note that this framework of utilizing diffusion models in adversarial training is generalizable. This raises the question of whether introducing this framework into GANs could improve the quality and diversity of samples generated by adversarial networks.

\subsection{E \ \ \ Compared with sota on more trajectories}\label{Compared-with-sota-on-more-trajectories}
We compared our method with the currently state-of-the-art baselines on more trajectories, as shown at the top of Figures~\ref{4-trajectories} and ~\ref{16-trajectories}. It can be observed that the performance of all methods improves as the number of trajectories increases, with faster and more stable convergence. Valuedice utilizes distribution correction estimation for off-policy and even offline policy learning. Its performance on 16 trajectories surprised us, especially the extremely fast and stable convergence on the walker2d task. However, Valuedice requires full trajectory demonstrations, which is a drawback. Our method also exhibited strong performance, achieving expert-level scores across all tasks and significantly surpassing the expert demonstrations substantially on the HalfCheetah and Ant tasks.

Additionally, we conducted multi-trajectory experiments under a state-only setting, with results shown at the bottom of Figures~\ref{4-trajectories} and ~\ref{16-trajectories}. Our method demonstrates solid performance, especially on the Ant and HalfCheetah tasks, substantially surpassing the expert demonstration performance. CFIL and OPOLO do not achieve this level of performance. 
\end{document}